# MRF-PINN: A Multi-Receptive-Field convolutional physics-informed neural network for solving partial differential equations

Shihong Zhang[a]   Chi Zhang[a]   Bosen Wang[a,*]

[a] *National Key Laboratory of Science and Technology on Aero-Engine Aero-thermodynamics, Research Institute of Aero-Engine, Beihang University, Beijing 100191, China*

**Abstract**

Compared with conventional numerical approaches to solving partial differential equations (PDEs), physics-informed neural networks (PINN) have manifested the capability to save development effort and computational cost, especially in scenarios of reconstructing the physics field and solving the inverse problem. Considering the advantages of parameter sharing, spatial feature extraction and low inference cost, convolutional neural networks (CNN) are increasingly used in PINN. However, some challenges still remain as follows. There are few studies on the general convolutional PINN scheme. To adapt convolutional PINN to solve different PDEs, considerable effort is usually needed for tuning critical hyperparameters. Furthermore, the effects of the finite difference accuracy, the model complexity, and the mesh resolution on the predictivity of convolutional PINN are not settled. To fill the gaps above, we propose three initiatives in this paper: (1) A Multi-Receptive-Field PINN (MRF-PINN) model is established to solve different types of PDEs on various mesh resolutions without manual tuning; (2) The dimensional balance method is used to estimate the loss weights when solving Navier-Stokes equations; (3) The Taylor polynomial is used to pad the virtual nodes near the boundaries for implementing high-order finite difference. The proposed MRF-PINN is tested for solving three typical linear PDEs (elliptic, parabolic, hyperbolic) and a series of nonlinear PDEs (Navier-Stokes equations) to demonstrate its generality and superiority. This paper shows that MRF-PINN can adapt to completely different equation types and mesh resolutions without any hyperparameter tuning. The dimensional balance method saves computational time and improves the convergence for solving Navier-Stokes equations. Further, the solving error is significantly decreased under high-order finite difference, large channel number, and high mesh resolution, which is expected to be a general convolutional PINN scheme.

---

\* Corresponding author at: National Key Laboratory of Science and Technology on Aero-Engine Aero-thermodynamics, Research Institute of Aero-Engine, Beihang University, Beijing 100191, China
E-mail address: wangbosen@buaa.edu.cn (B.S. Wang).

# 1. Introduction

The conventional numerical methods for solving partial differential equations (PDEs) in various scenarios such as fluid mechanics[1], solid mechanics[2], heat transfer[3], and electromagnetism[4] based on numerical discretisation have vigorously developed in recent decades. The representative methods include the finite difference[5], the finite element[6], the finite volume[7], the Lattice Boltzmann[8], and the smooth particle method[9]. Integrating the available data of high fidelity is promising to improve accuracy and efficiency for solving PDEs, however, it is quite challenging for the conventional numerical methods to achieve. In addition, the considerable computational cost for solving the inverse problem [10] is a long-term standing issue. In recent years, physics-informed neural networks (PINN) have been developed as an alternative to overcome the challenges above. PINN can also eliminate the dependence on large dataset by incorporating the constraints of prior physical principles[11][12][13][14][15]. The typical neural network architectures embedded in PINN are multilayer perceptron (MLP) and convolutional neural network (CNN). Due to the advantages of parameter sharing, spatial feature extraction, and lightweight, CNN is progressively becoming favoured [16][17][18]. However, the development of convolutional PINN is still at a relatively preliminary stage and has plenty of scope for improvement of its convergence rate and accuracy for solving forward and inverse problems[14], especially in terms of model architectures [19], weight estimation [20] and finite difference accuracy [16].

Researchers are committed to improving the efficiency, accuracy, and generality of PINN in solving forward and inverse problems from series aspects. With regard to the model architecture, ResNet, U-Net, and Fourier neural operator are usually used to build PINN models [19][21][22], which can inhibit the model degradation, adapt to the image-to-image prediction, and accelerate the convergence. Concerning the activation function, Jagtap *etal.* proposed one with adaptive parameters for PINN which can accelerate the speed of convergence and reduce the solving error [23]. In respect of the multi-fidelity data integration, Meng *etal.* used a composite PINN to improve the model accuracy meanwhile weakening dependence on high-cost and high-fidelity data [24]. Regarding the computational domain transformation, Gao *etal.* transformed an irregular complex domain into a regular rectangular one through coordinate transformation in order to realise the deployment and gradient computations in CNN for complex computational geometries [25]. In terms of sampling method and physical constraint, LuLu *etal.* adapted residual-based adaptive refinement and gradient-enhanced, which can reduce the number of sampling points, speed up the model training process, and reduce the solving error [26]. To estimate reasonable loss weights, Wang *etal.* proposed a dynamic weight estimation method to balance multiple loss functions in PINN [20]. For the choice of the optimization method, Adam and L-BFGS are usually used to optimize and fine-tune the model, respectively [13].

The methods as mentioned above significantly improve the generality and the accuracy of PINN. However, for convolutional PINN systematic investigations on the general network model, the estimation of loss weights, the finite difference accuracy, the model complexity, and the mesh resolution are far from sufficient. This leads to obstacles for CNN to present its advantages of the efficient feature extraction and the lightweight inference in PINN. Accordingly, we develop a convolutional PINN model with multiple receptive fields (MRF-PINN). Here, MRF-PINN can be adapted to solving different type of PDEs without manual tuning which is expected to ensure the generality of the convolutional PINN. Further numerical verification also proves the advantages of the MRF-PINN scheme in prediction accuracy and convergence.

This paper will be organised as follows: In Section 2, the algorithm of the multiple receptive fields concept for PINN will be described which is followed by the associated methodology of boundary padding, higher-order finite difference, loss functions, and weight estimation. In Section 3, the numerical details of MRF-PINN for solving three linear PDEs (elliptic, parabolic, hyperbolic) and one series of nonlinear PDEs (Navier-Stokes PDEs) are presented, and the results under different mesh resolutions are compared with those from FEM and FVM to corroborate its validity. In Section 4, the superiority in generality of MRF-PINN is verified by comparing with the traditional fixed receptive field PINN. The dimension balance method as proposed is demontrated to be more effective than the dynamic weight method based on the tests on solving the Navier-Stokes PDEs. Moreover, the effect of the finite difference accuracy on the performance of MRF-PINN are tested systematically.

## 2. Methodology

According to the illustration of the MRF-PINN as shown in Fig. 1, its algorithm is elaborated as follows:

---

MRF-PINN algorithm

---

(a) Initialise the input via mean filtering.

(b) Estimate the loss weights $\lambda_{\text{PDEs}}$, $\lambda_{\text{BC}}$, $\lambda_{\text{IC}}$, $\lambda_{\text{data}}$.

**for** epoch=1, 2, 3, ..., epoch$_{\text{Adam}}$ **do**

    (c) Perform forward-propagation for six encoder-decoder pairs.

    (d) Recombine the six intermediate physics fields output by encoder-decoders, and place the same components together.

    (e) Obtain the result of MRF-PINN by linearly superimposing the recombined physical fields.

(f) Calculate the mean square error (MSE) loss of the PDEs $L_{\text{PDEs}}$.

(g) Calculate the MSE loss of the boundary conditions $L_{\text{BC}}$ and initial conditions $L_{\text{IC}}$.

(h) Calculate the MSE loss of the known data $L_{\text{data}}$.

(i) Assigne $L_{\text{PDEs}}$, $L_{\text{BC}}$, $L_{\text{IC}}$, $L_{\text{data}}$ with the loss weights $\lambda_{\text{PDEs}}$, $\lambda_{\text{BC}}$, $\lambda_{\text{IC}}$, $\lambda_{\text{data}}$, respectively, and sum up as the total loss $L$.

(j) Optimise the parameters of six encoder-decoder pairs and the linear superposition operation by using the Adam optimization algorithm, with a learning rate of 1e-4.

**end**

**for** epoch=1, ..., epoch$_{\text{L-BFGS}}$ **do**

(k) The same steps as (c)~(i).

(l) Fine-tune the parameters by using the L-BFGS optimisation algorithm.

**end**

(m) Output the final MRF-PINN result.

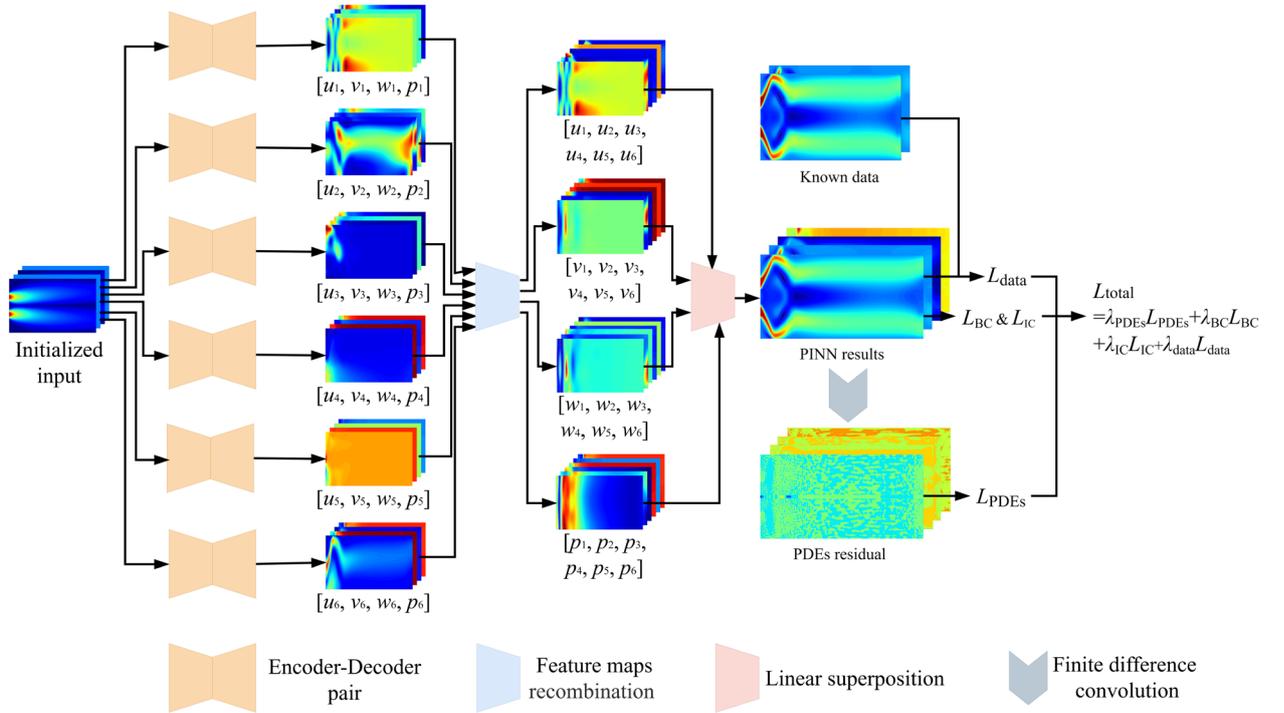

Fig. 1. MRF-PINN Model

## 2.1 MRF-PINN model

For convolutional PINN, if the mesh resolution of the input and output changes, the receptive field of the convolution kernel relative to the physics will also change simultaneously. For example, increasing the mesh resolution when solving the Navier-Stokes PDEs, the flow field structure will be described by more mesh nodes. If the size of the convolution kernel remains unchanged, its receptive field will be reduced relative to the flow field structure. The consequences are as follows: (1) The hyperparameters such as the optimal receptive field need to be determined by duplicated manual tuning. (2) At different mesh resolutions, the model's generalization will be weakened because the receptive field of the convolution kernel relative to the physical field has significantly changed, and the nonlinear mapping relationship between input and output is seriously invalid.

As shown in Fig. 2, a resolution-adapted multiple receptive field dilated convolution is proposed in this paper. A corresponding Encoder-Decoder pair is built as shown in Fig. 3.

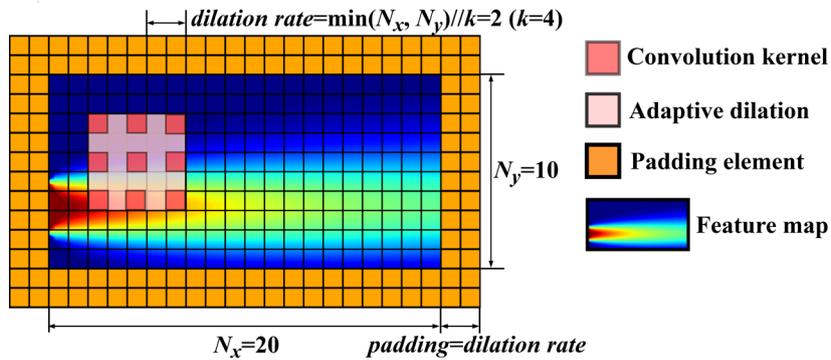

Fig. 2. Resolution-adapted dilated convolution, The size of the receptive field is determined by the dilation rate of the convolution kernel (The receptive field of 3×3 is padded to 5×5 with dilation rate=2), and the dilation rate of the convolution kernel is set to a fixed multiple of the input resolution (the multiple in the figure is 1/4).

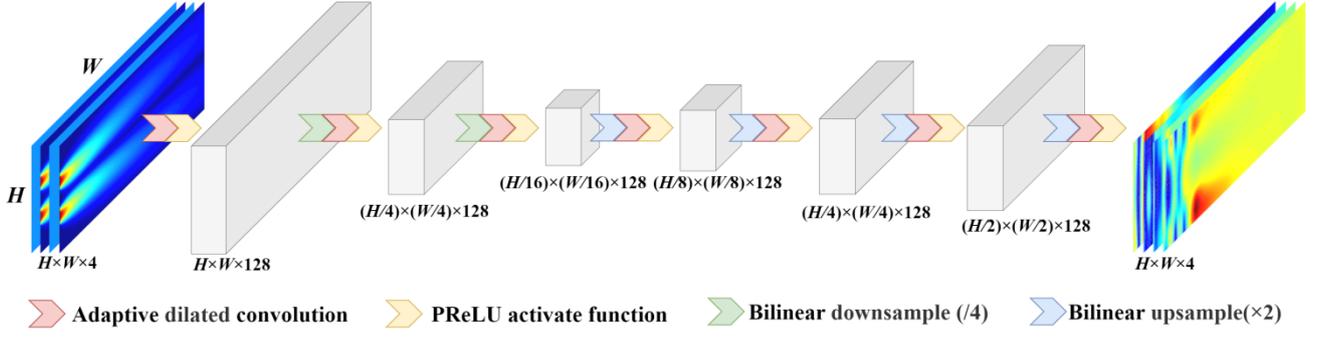

Fig. 3. Single Encoder-Decoder pair

In Fig. 2, the dilation rate of the convolution kernel and the padding number are both dependent on the size of the feature map. As illustrated in Fig. 3, in one layer of the Encoder-Decoder pair, the size of feature map is $c \times h \times w$ where $c$ represents the channel number of the input feature map, and $h \times w$ is the size of each channel. The dilated rate of the convolution kernel on this layer and the padding number of the feature map are determined to be $\max(\lfloor \min(h,w)//k \rfloor, 1)$. And $k$ denotes the ratio of feature map size and receptive field size, which are set to be 2, 4, 8, 16, 32, 64 in six Encoder-Decoder pairs, respectively. In this way, when each Encoder-Decoder propagates forward, the receptive field of each convolution kernel remains unchanged relative to its feature map, which is always $2 \times max(\lfloor \min(h,w), 1 \rfloor) + 1$. The receptive fields of different Encoder-Decoder pairs are different, and the receptive fields of the six Encoder-Decoder pairs decrease relative to the feature maps. Finally, the outputs of the six Encoder-Decoder pairs are linearly superimposed to obtain the final prediction of MRF-PINN which is given by

$$u_i = \text{Encoder-Decoder}_i(u_{initial}) \quad i = 1,2,3,4,5,6 \qquad (1)$$

$$u_{PINN} = \sum_{i=1}^{6} \theta_i \cdot u_i = \sum_{i=1}^{6} \theta_i \cdot \text{Encoder-Decoder}_i(u_{initial}) \qquad (2)$$

**2.2 Boundary padding and high-order finite difference**

Convolutional PINN performs finite difference through a fixed-parameter convolution kernel, and the accuracy of the finite difference directly determines the precision of the derivative.

Before performing finite-difference convolution, the boundary of the PINN result should be reasonably padded or one-sided difference near the boundary. In this paper, a Taylor polynomial fitting is used to pad the boundary which can ensure accuracy and concision of the computations of finite difference near the boundary.

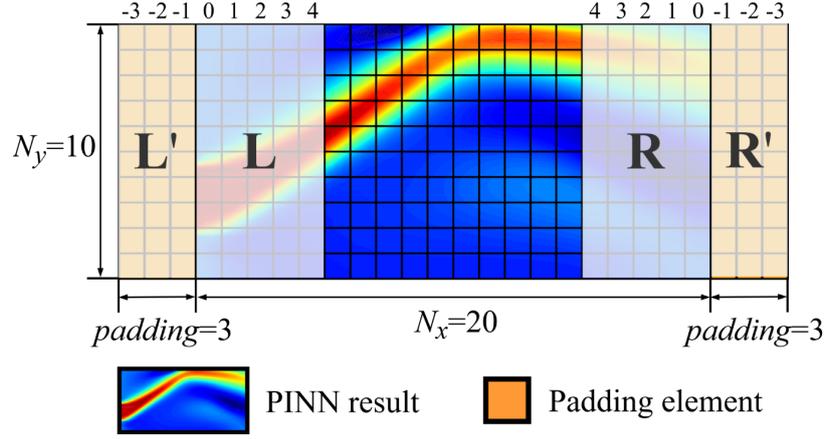

Fig. 4. Boundary padding for a six-order finite difference convolution

The boundary padding for a sixth-order central difference is taken as an example and is illustrated in Fig. 4. Assume that the variable $u$ and its first three partial derivatives at $n$ ($n$=-3, -2, -1, 0, 1, 2, 3) nodes away from the boundary of PINN result are $u^{(n)}, u_x^{(n)}, u_{xx}^{(n)}, u_{xxx}^{(n)}$, respectively. According to Taylor expansion:

$$u^{(n)} = u^{(0)} + n \cdot u_x^{(0)} + \frac{n^2}{2} \cdot u_{xx}^{(0)} + \frac{n^3}{6} \cdot u_{xxx}^{(0)} + o(n^3) \quad n = -3, -2, -1, 0, 1, 2, 3 \quad (3)$$

$$\begin{bmatrix} u^{(1)} - u^{(0)} \\ u^{(2)} - u^{(0)} \\ u^{(3)} - u^{(0)} \end{bmatrix} = \begin{bmatrix} 1 & 1/2 & 1/6 \\ 1 & 4/2 & 8/6 \\ 1 & 9/2 & 27/6 \end{bmatrix} \begin{bmatrix} u_x^{(0)} \\ u_{xx}^{(0)} \\ u_{xxx}^{(0)} \end{bmatrix} + o(n^3) \quad (4)$$

For the padded virtual nodes, we can have:

$$\begin{bmatrix} u^{(-1)} - u^{(0)} \\ u^{(-2)} - u^{(0)} \\ u^{(-3)} - u^{(0)} \end{bmatrix} = \begin{bmatrix} -1 & 1/2 & -\dfrac{1}{6} \\ -1 & 4/2 & -\dfrac{8}{6} \\ -1 & 9/2 & -\dfrac{27}{6} \end{bmatrix} \begin{bmatrix} u_x^{(0)} \\ u_{xx}^{(0)} \\ u_{xxx}^{(0)} \end{bmatrix} + o(n^3) \qquad (5)$$

The values at the padded virtual nodes can be obtained as

$$\begin{bmatrix} u^{(-1)} \\ u^{(-2)} \\ u^{(-3)} \end{bmatrix} \approx \begin{bmatrix} u^{(0)} \\ u^{(0)} \\ u^{(0)} \end{bmatrix} + \begin{bmatrix} -1 & 1/2 & -\dfrac{1}{6} \\ -2 & 4/2 & -\dfrac{8}{6} \\ -3 & 9/2 & -\dfrac{27}{6} \end{bmatrix} \begin{bmatrix} 1 & 1/2 & 1/6 \\ 2 & 4/2 & 8/6 \\ 3 & 9/2 & 27/6 \end{bmatrix}^{-1} \begin{bmatrix} u^{(1)} - u^{(0)} \\ u^{(2)} - u^{(0)} \\ u^{(3)} - u^{(0)} \end{bmatrix} \qquad (6)$$

In most cases, the distribution of physical variable $u$ near the boundary is relatively smooth, and the size of the finite difference operator is generally much smaller than the total number of nodes in the derivation direction. Therefore, it can also be assumed that in the derivation direction near the boundary, the distribution of physical variable $u$ is a smooth cubic distribution (or quadratic distribution), that is:

$$u^{(n)} \equiv u^{(0)} + n \cdot u_x^{(0)} + \frac{n^2}{2} \cdot u_{xx}^{(0)} + \frac{n^3}{6} \cdot u_{xxx}^{(0)} \qquad n = -3, -2, -1, 0, 1, 2, 3 \qquad (7)$$

Obviously, this formula is consistent with the third-order Taylor expansion. Both methods can be used to determine the value of the padded virtual nodes. In order to avoid being affected by some outliers, more internal nodes can be used to determine the value of the padded virtual nodes. But the equation is over-determined, and the pseudo-inverse can be used instead of the original inverse which is given by:

$$\begin{bmatrix} u^{(-1)} \\ u^{(-2)} \\ u^{(-3)} \end{bmatrix} \approx \begin{bmatrix} u^{(0)} \\ u^{(0)} \\ u^{(0)} \end{bmatrix} + \begin{bmatrix} -1 & 1/2 & -\dfrac{1}{6} \\ -2 & 4/2 & -\dfrac{8}{6} \\ -3 & 9/2 & -\dfrac{27}{6} \end{bmatrix} \begin{bmatrix} 1 & 1/2 & 1/6 \\ 2 & 4/2 & 8/6 \\ 3 & 9/2 & 27/6 \\ \cdots \end{bmatrix}^{+} \begin{bmatrix} u^{(1)} - u^{(0)} \\ u^{(2)} - u^{(0)} \\ u^{(3)} - u^{(0)} \\ \cdots \end{bmatrix} \qquad (8)$$

By boundary padding, the finite difference convolution can be performed to calculate the PDE loss. The finite difference convolution based on a six-order central difference is shown in Fig. 5. The number of the padded virtual nodes is half of the central difference order, which ensures the preservation of the size of the PINN result and the derivative result through the central difference convolution. This paper adopts 2-order Taylor polynomials for boundary padding to robustly preserve the accuracy of derivatives on the boundary since the highest order of PDE is 2. Then, the finite difference convolution kernel with fixed parameters is imposed on the PINN result to obtain the derivative value of the physical field. The parameters of the convolution kernel for calculating the first and the second derivatives are shown in Table 1:

Table 1

Parameters of the finite difference convolution kernel

| Derivative order | Finite difference order | Parameters of the convolution kernel |
|---|---|---|
| 1 | 2 | [-1/2, 0 ,1/2] |
|  | 4 | [1/12, -2/3, 0, 2/3, -1/12] |
|  | 6 | [-1/60, 3/20, -3/4, 0, 3/4, -3/20, 1/60] |
|  | 8 | [1/280, -4/105, 1/5, -4/5, 0, 4/5, -1/5, 4/105, -1/280] |
| 2 | 2 | [1, -2, 1] |
|  | 4 | [-1/12, 4/3, -5/2, 4/3, -1/12] |
|  | 6 | [1/90, -3/20, 3/2, -49/18, 3/2, -3/20, 1/90] |
|  | 8 | [-1/560, 8/315, -1/5, 8/5, -205/72, 8/5, -1/5, 8/315, -1/560] |

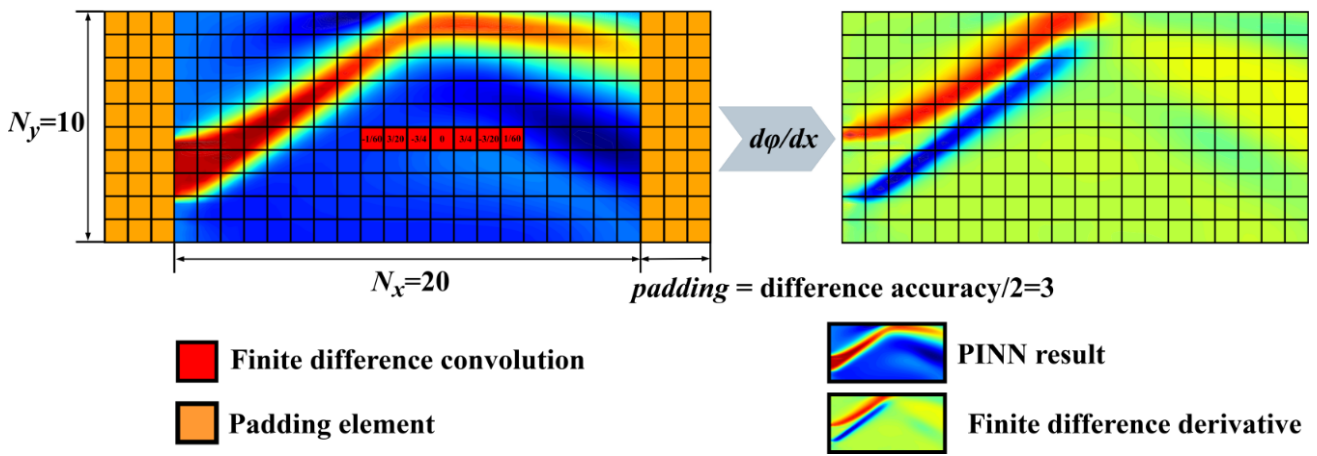

Fig. 5. Boundary padding and six-order finite difference convolution

## 2.3 Loss functions and weight estimation

The prediction of PINN is compelled to satisfy the constraints of PDEs, boundary condition, initial condition and known data. These terms are considered by a composed loss function, a weighted sum of PDE loss, boundary loss, initial loss and data loss. Thus, the loss weight will directly impact the convergence and solving error of PINN. The dimensional balance method proposed here is expected to be an informative weight estimation method.

For a PINN network $\hat{y} = \text{PINN}(x;\theta)$, its loss function consists of the following parts:

(1) PDE loss. Variables $x$ and $y$ should satisfy the PDE relationship $f\left(x, y, \frac{dy}{dx}, \frac{d^2y}{dx^2}, \cdots\right) = 0$. Therefore, the PDE MSE loss is defined as

$$L_{PDEs} = \frac{1}{N_\Omega} \sum_i f^2\left(x_i, y_i, \frac{dy_i}{dx_i}, \frac{d^2y_i}{dx_i^2}, \cdots\right) \tag{9}$$

$N_\Omega$ is the number of the internal nodes.

(2) Boundary condition loss, which is the MSE error between the predicted boundary value and the known boundary condition. For a Dirichlet type boundary field, the boundary condition loss is determined as

$$L_{BC} = \frac{1}{N_{\partial\Omega}} \sum_i (\hat{y}_i - y_i)^2 \tag{10}$$

$N_{\partial\Omega}$ is the number of the boundary nodes.

(3) Initial condition loss, which is the MSE error between the predicted value $\hat{y}$ at $t=0$ and the given initial condition. The MSE loss of the initial condition is calculated as

$$L_{IC} = \frac{1}{N_{t=0}} \sum_i (\hat{y}_i - y_i)^2 \tag{11}$$

$N_{t=0}$ is the number of the initial nodes.

(4) Data loss, which is the MSE error between the PINN prediction $\hat{y}$ and the known data. The MSE loss of the data is expressed as

$$L_{data} = \frac{1}{N_{data}} \sum_i (\hat{y}_i - y_i)^2 \qquad (12)$$

$N_{data}$ is the number of nodes whose value is known.

The total loss of PINN is the weighted sum of the losses above which is

$$L = \lambda_{PDEs} L_{PDEs} + \lambda_{BC} L_{BC} + \lambda_{IC} L_{IC} + \lambda_{data} L_{data} \qquad (13)$$

The back-propagation algorithm can be used to optimise the parameters $\theta$ in $\hat{y} = NN(x; \theta)$ until the loss meets the criterion for convergence. The loss weights $\lambda_{PDEs}, \lambda_{BC}, \lambda_{IC}, \lambda_{data}$ directly determine the influence of each loss on the parameter optimisation and in consequence affects the convergence of PINN.

In most of the existing research, the loss weights are usually determined by manual tuning or dynamic weight estimation. Besides, this paper proposes a dimensional balance method for the Navier-Stokes PDEs.

Manual tuning is a tedious and time-consuming task, which is only operable when the number of loss functions is few. In this paper, there are only PDE loss $L_{PDEs}$ and boundary condition loss $L_{BC}$ when solving the linear PDEs, so their loss weights are set manually.

The dynamic weight method utilizes the gradient information of each loss to the parameters, then balances the dependence on each part of the loss when parameters are updated. The dynamic weight method eliminates the ambiguousness of human tuning while it may lead to more back-propagation and thereby the deceleration of optimisation and convergence. The dynamic weight method is executed by following steps：

(1) Back-propagation is performed for the PDE loss $L_{PDEs}$, the boundary condition loss $L_{BC}$, the initial condition loss $L_{IC}$, the data loss $L_{data}$ to obtain $\nabla_\theta L_{PDEs}, \nabla_\theta L_{BC}, \nabla_\theta L_{IC}, \nabla_\theta L_{data}$ and $\overline{|\nabla_\theta L_{PDEs}|}, \overline{|\nabla_\theta L_{BC}|}, \overline{|\nabla_\theta L_{IC}|}, \overline{|\nabla_\theta L_{data}|}$, respectively.

(2) Calculate the new loss weights:

$$\hat{\lambda}_{PDEs}^{(k+1)} = 1 \qquad (14a)$$

$$\hat{\lambda}_{BC}^{(k+1)} = \overline{|\nabla_\theta L_{PDEs}|}/\overline{|\nabla_\theta L_{BC}|} \tag{14b}$$

$$\hat{\lambda}_{IC}^{(k+1)} = \overline{|\nabla_\theta L_{PDEs}|}/\overline{|\nabla_\theta L_{IC}|} \tag{14c}$$

$$\hat{\lambda}_{data}^{(k+1)} = \overline{|\nabla_\theta L_{PDEs}|}/\overline{|\nabla_\theta L_{data}|} \tag{14d}$$

(3) Update the loss weights, $\alpha=0.1$ is the update weight coefficient.

$$\lambda_{PDEs}^{(k+1)} = 1 \tag{15a}$$

$$\lambda_{BC}^{(k+1)} = \alpha\hat{\lambda}_{BC}^{(k+1)} + (1-\alpha)\lambda_{BC}^{(k)} \tag{15b}$$

$$\lambda_{IC}^{(k+1)} = \alpha\hat{\lambda}_{IC}^{(k+1)} + (1-\alpha)\lambda_{IC}^{(k)} \tag{15c}$$

$$\lambda_{data}^{(k+1)} = \alpha\hat{\lambda}_{data}^{(k+1)} + (1-\alpha)\lambda_{data}^{(k)} \tag{15d}$$

This paper uses the dynamic weight method to assist in setting the loss weights in solving linear PDEs. However, the dynamic weight method has been verified invalid when solving the Navier-Stokes PDEs due to abundant loss functions and the magnitude difference. However, the dimensions of the Navier-Stokes PDEs are easy to estimate, and the dimensional estimation method has been maturely applied to the study of fluids. Therefore, in this paper the loss weights are determined by a dimensional balance approach:

(1) Normalise the relevant variables to $\tilde{x}$, $\tilde{y}$, $\tilde{\tilde{y}}$ and thus to separate the dimension of each loss function:

$$L_{PDEs} = \frac{1}{N_\Omega}\sum_i f^2\left(x_i, y_i, \frac{dy_i}{dx_i}, \frac{d^2y_i}{dx_i^2}, \cdots\right) = X^a Y^b \cdot \frac{1}{N_\Omega}\sum_i f^2\left(\tilde{x}_i, \tilde{y}_i, \frac{d\tilde{y}_i}{d\tilde{x}_i}, \frac{d^2\tilde{y}_i}{d\tilde{x}_i^2}, \cdots\right) \tag{16a}$$

$$L_{BC} = X^c Y^d \frac{1}{N_{\partial\Omega}}\sum_i \left(\tilde{\tilde{y}}_i - \tilde{y}_i\right)^2 \tag{16b}$$

$$L_{IC} = X^e X^f \frac{1}{N_{t=0}}\sum_i \left(\tilde{\tilde{y}}_i - \tilde{y}_i\right)^2 \tag{16c}$$

$$L_{data} = X^g X^h \frac{1}{N_{data}}\sum_i \left(\tilde{\tilde{y}}_i - \tilde{y}_i\right)^2 \tag{16d}$$

(2) Perform dimensional balance (or normalisation) for each loss function:

$$\tilde{L}_{PDEs} = X^{-a} Y^{-b} \cdot L_{PDEs} = \frac{1}{N_\Omega}\sum_i f^2\left(\tilde{x}_i, \tilde{y}_i, \frac{d\tilde{y}_i}{d\tilde{x}_i}, \frac{d^2\tilde{y}_i}{d\tilde{x}_i^2}, \cdots\right) \tag{17a}$$

$$\tilde{L}_{BC} = X^{-c}Y^{-d} \cdot L_{BC} = \frac{1}{N_{\partial\Omega}}\sum_i \left(\tilde{\hat{y}}_i - \tilde{y}_i\right)^2 \tag{17b}$$

$$\tilde{L}_{IC} = X^{-e}X^{-f} \cdot L_{IC} = \frac{1}{N_{t=0}}\sum_i \left(\tilde{\hat{y}}_i - \tilde{y}_i\right)^2 \tag{17c}$$

$$\tilde{L}_{data} = X^{-g}X^{-h} \cdot L_{data} = \frac{1}{N_{data}}\sum_i \left(\tilde{\hat{y}}_i - \tilde{y}_i\right)^2 \tag{17d}$$

(3) Calculate the total loss as

$$\begin{aligned} L &= \tilde{L}_{PDEs} + \tilde{L}_{BC} + \tilde{L}_{IC} + \tilde{L}_{data} \\ &= X^{-a}Y^{-b} \cdot L_{PDEs} + X^{-c}Y^{-d} \cdot L_{BC} + X^{-e}Y^{-f} \cdot L_{IC} + X^{-g}Y^{-h} \cdot L_{data} \end{aligned} \tag{18}$$

and the loss weights are obtained as

$$\lambda_{PDEs} = X^{-a}Y^{-b} \tag{19a}$$

$$\lambda_{BC} = X^{-c}Y^{-d} \tag{19b}$$

$$\lambda_{IC} = X^{-e}Y^{-f} \tag{19c}$$

$$\lambda_{data} = X^{-g}Y^{-h} \tag{19d}$$

## 3. Setting and validation

The proposed MRF-PINN, the finite element method (FEM), and the finite volume method (FVM) are used to solve three types of linear PDEs (elliptic, hyperbolic, parabolic) and one series of nonlinear PDEs (Navier-Stokes PDEs). These PDEs have been widely used for test and validation of the PINN models[13][14][25]. To quantify the solving error of MRF-PINN, we define the two-norm error $\varepsilon$ representing the difference from the MRF-PINN result to the FEM / FVM result, which are given by

$$\varepsilon = \|u_{PINN} - u_{FEM}\|_2 / \|u_{FEM}\|_2 \tag{20a}$$

or

$$\varepsilon = \|u_{PINN} - u_{FVM}\|_2 / \|u_{FVM}\|_2 \tag{20b}$$

Based on the numerical test, the MRF-PINN model is validated and its advantages are verified. The FEM and FVM results are considered absolutely correct because they are solved from the fine mesh resolution and verified independent of the finer mesh resolution.

## 3.1 Linear PDEs

Second-order linear PDEs can be used to describe numerous physical phenomena precisely and concisely, such as heat conduction, electromagnetic, gravity, diffusion, and vibration. Second-order linear PDEs can be typically categorised into elliptic, parabolic, and hyperbolic equations. The dependence and influence domains of these three kinds of equations are different, so they can describe various physical phenomena of equilibrium, approaching equilibrium, and non-equilibrium.

A general second-order linear PDE can be expressed as

$$a\frac{\partial^2 u}{\partial x^2} + b\frac{\partial^2 u}{\partial x \partial y} + c\frac{\partial^2 u}{\partial y^2} + d\frac{\partial u}{\partial x} + e\frac{\partial u}{\partial y} + fu = g \tag{21}$$

Elliptic ($b^2 - 4ac < 0$), parabolic ($b^2 - 4ac = 0$), and hyperbolic ($b^2 - 4ac > 0$) PDEs can be obtained by adjusting the corresponding coefficients. The PDE coefficients for the present test are listed in Table 2.

Table 2

The setup of the second-order linear PDEs

| PDE types | a | b | c | d | e | f | g |
|---|---|---|---|---|---|---|---|
| Elliptic | 1 | 0 | 1 | 2 | 2 | 4 | 4 |
| Parabolic | 1 | 0 | 0 | -2 | -2 | 4 | 4 |
| Hyperbolic | 1 | 0 | -1 | 0 | 0 | 0 | 0 |

The numerical setup of the MRF-PINN is briefly introduced as follows. The mesh resolution of input and output is $32 \times 64$. The input is the initial field which is filtered from the boundary value

$$I(n+1) = I(n) \otimes [1/3 \quad 1/3 \quad 1/3] \quad n = 0,1,2,\cdots N_x (\text{or } N_y) \tag{22}$$

where *n* is the distance from the boundary to an internal row or column. The receptive field sizes of the six Encoder-Decoder pairs are 33, 17, 9, 5, 3, and 3, respectively, which are dilated by a $3 \times 3$ convolution kernel. The relevant parameters for the dilatation are shown in Table 3. The padded virtual nodes are obtained by fitting a 2-order Taylor polynomial. An 8-order central difference is employed for discretisation. The total loss is composed of the PDE loss and the boundary condition loss, the loss weights are set to be 1 and 1000. During the solving process, the Adam optimiser is used to optimise 9000 epochs with a learning rate of 1e-4. The L-BFGS optimiser is set to run 500 epochs for fine-tuning MRF-PINN.

Table 3

Details of MRF-PINN dilated convolution for solving linear PDEs.

| Enc-Dec | Enc-Dec-1 | Enc-Dec-2 | Enc-Dec-3 | Enc-Dec-4 | Enc-Dec-5 | Enc-Dec-6 |
|---|---|---|---|---|---|---|
| $k$ | 2 | 4 | 8 | 16 | 32 | 64 |
| Dilation rate | 16 | 8 | 4 | 2 | 1 | 1 |
| Padding | 16 | 8 | 4 | 2 | 1 | 1 |
| Receptive field | 33 | 17 | 9 | 5 | 3 | 3 |
| Receptive field/min($h,w$) | 33/32≈1.03 | 17/32≈0.53 | 9/32≈0.28 | 5/32≈0.16 | 3/32≈0.09 | 3/32≈0.09 |

The FEM simulation is conducted by using the solver MUMPS on a mesh resolution of $200 \times 400$. In the MUMPS numerical solver, the equations and constraints are discretized on the elements to solve the linear equation $\mathbf{A}x = b$. MUMPS solves the inverse matrix of $\mathbf{A}$ to solve the PDE, and its realization method is Gaussian decomposition.

### 3.1.1 Elliptic PDE

The elliptic PDEs are usually used to describe equilibrium states, such as steady temperature fields and steady potential distributions. An elliptic PDE does not have characteristic lines. Its dependence and influence domain covers the entire domain.

The elliptic PDE and its boundary conditions for the current test are

$$\begin{cases} \dfrac{\partial^2 u}{\partial x^2} + \dfrac{\partial^2 u}{\partial y^2} + 2\dfrac{\partial u}{\partial x} + 2\dfrac{\partial u}{\partial y} + 4u = 4 & 0 < x < 1, 0 < y < 2 \\ u|_{x=0} = -\sin(\pi y) \quad u|_{x=1} = \sin(\pi y) & 0 < y < 2 \\ u|_{y=0} = \sin(2\pi x) \quad u|_{y=2} = -\sin(2\pi x) & 0 < x < 1 \end{cases} \quad (23)$$

Take an example and as shown in Fig. 6, MRF-PINN can reproduce the results from FEM well where the solving error $\varepsilon$ is 0.0127.

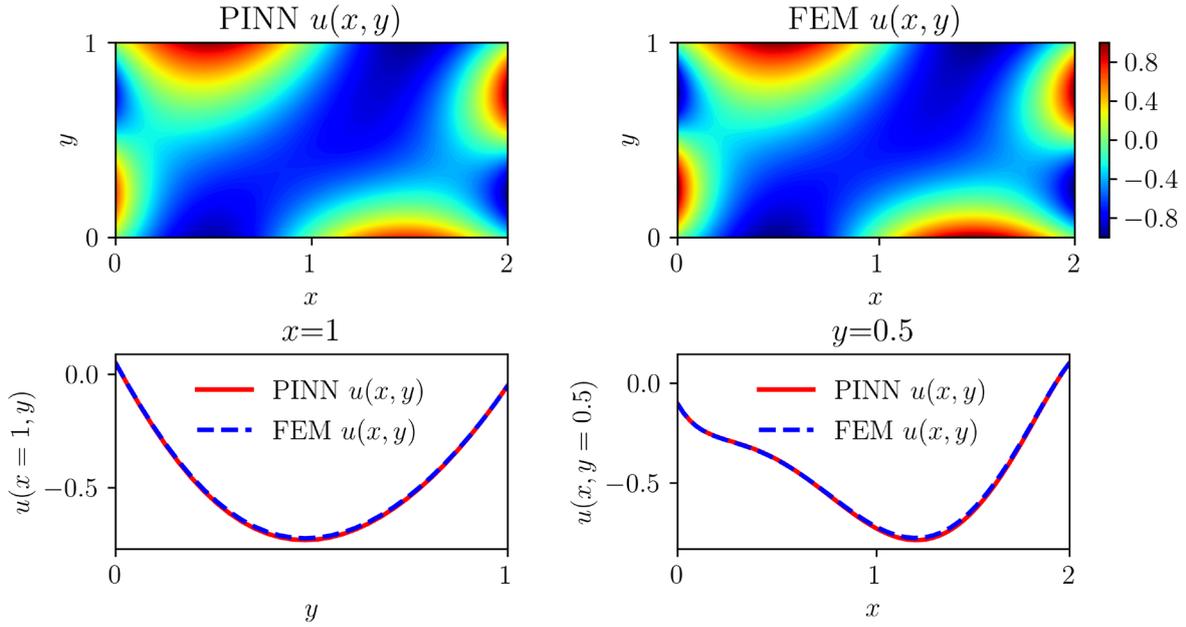

Fig. 6. Solutions of the elliptic PDE by MRF-PINN and FEM

### 3.1.2 Parabolic PDE

Parabolic PDEs are typical expressions for elucidating the process to approach a steady state, such as heat conduction and diffusion. A parabolic PDE has only one characteristic line such that one state only depends on the previous moment. The parabolic PDE and its boundary conditions for test are expressed as

$$\begin{cases} \dfrac{\partial^2 u}{\partial x^2} - 2\dfrac{\partial u}{\partial x} - 2\dfrac{\partial u}{\partial y} + 4u = 4 & 0 < x < 1, 0 < y < 2 \\ u|_{x=0} = -\sin(\pi y) \quad u|_{x=1} = \sin(\pi y) & 0 < y < 2 \\ u|_{y=0} = \sin(2\pi x) & 0 < x < 1 \end{cases} \quad (24)$$

To visualize the results, Fig. 7 shows that the solutions of MRF-PINN and FEM

are in good agreement and the solving error $\varepsilon$ of MRF-PINN is 0.00736.

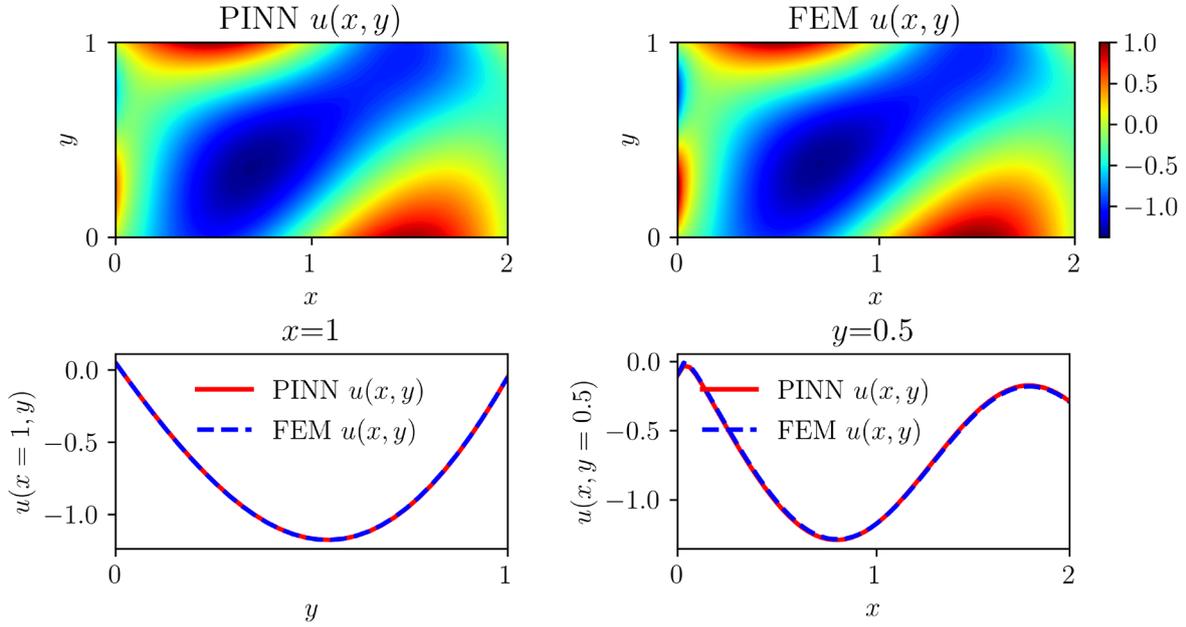

Fig. 7. Solutions of the parabolic PDE by MRF-PINN and FEM

### 3.1.3 Hyperbolic PDE

Hyperbolic PDEs are often used to model non-equilibrium wave phenomena, such as mechanical, sound, and electromagnetic waves. The hyperbolic PDE has two characteristic lines, hence the dependence and influence domain of one point in time and space is limited. The hyperbolic PDE and its boundary conditions tested here are

$$\begin{cases} \dfrac{\partial^2 u}{\partial x^2} - \dfrac{\partial^2 u}{\partial y^2} = 0 & 0 < x < 1, 0 < y < 2 \\ u|_{x=0} = 0 \quad u|_{x=1} = 0 & 0 < y < 2 \\ u|_{y=0} = \sin(2\pi x) \quad u|_{y=2} = \sin(2\pi x) & 0 < x < 1 \end{cases} \tag{25}$$

Fig. 8 gives an example of overall consistency between the results from PINN to FEM. A solving error $\varepsilon$ of 0.05512 is identified.

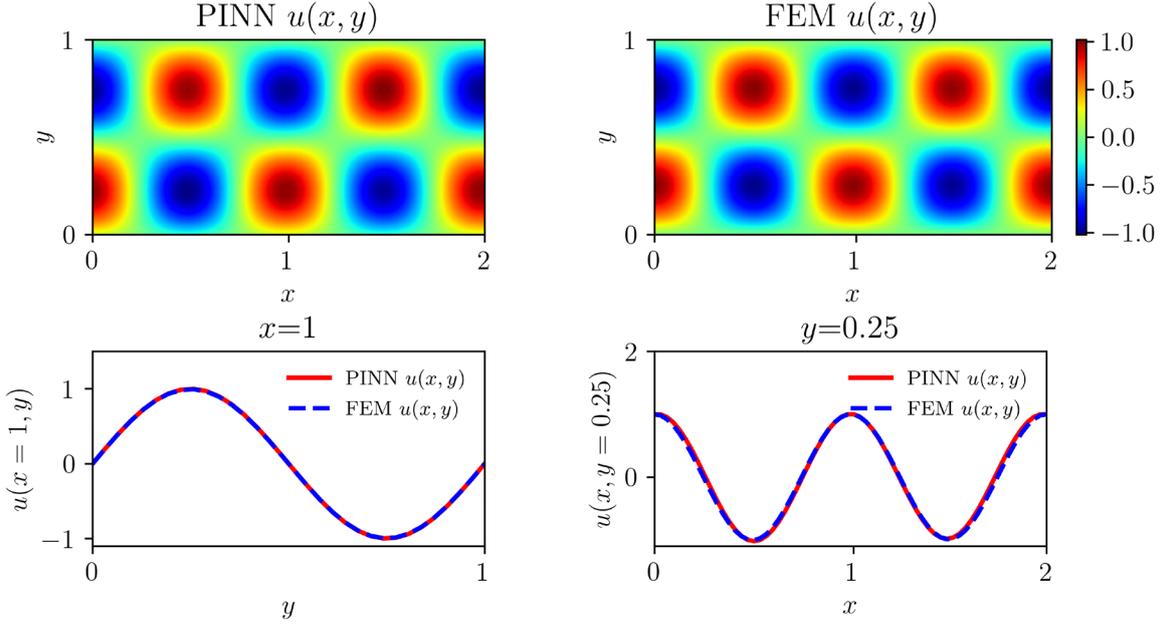

Fig. 8. Solutions of the hyperbolic PDE by MRF-PINN and FEM

### 3.2 Nonlinear PDEs: Navier-Stokes PDEs

The well-known Navier-Stokes PDEs are typical highly nonlinear PDEs for testing the validity of MRF-PINN. The Navier-Stokes PDEs solved in this paper is axisymmetric swirl flow in two-dimensional cylindrical coordinate.

$$\begin{cases} \frac{\partial u}{\partial z} + \frac{1}{r}\frac{\partial (rv)}{\partial r} = 0 & 0 < r < 1.5 \quad 0 < z < 18 \\ u\frac{\partial u}{\partial z} + v\frac{\partial u}{\partial r} = -\frac{1}{\rho}\frac{\partial p}{\partial x} + \nu\left(\frac{\partial^2 u}{\partial z^2} + \frac{\partial^2 u}{\partial r^2} + \frac{1}{r}\frac{\partial u}{\partial r}\right) & 0 < r < 1.5 \quad 0 < z < 18 \\ u\frac{\partial v}{\partial z} + v\frac{\partial v}{\partial r} - \frac{w^2}{r} = -\frac{1}{\rho}\frac{\partial p}{\partial r} + \nu\left(\frac{\partial^2 v}{\partial z^2} + \frac{\partial^2 v}{\partial r^2} + \frac{1}{r}\frac{\partial v}{\partial r} - \frac{v}{r^2}\right) & 0 < r < 1.5 \quad 0 < z < 18 \\ u\frac{\partial w}{\partial z} + v\frac{\partial w}{\partial r} - \frac{vw}{r} = \nu\left(\frac{\partial^2 w}{\partial z^2} + \frac{\partial^2 w}{\partial r^2} + \frac{1}{r}\frac{\partial w}{\partial r} - \frac{w}{r^2}\right) & 0 < r < 1.5 \quad 0 < z < 18 \\ v|_{r=0} = w|_{r=0} = 0, \quad \left.\frac{\partial u}{\partial r}\right|_{r=0} = 0 \quad u|_{r=1.5} = v|_{r=1.5} = w|_{r=1.5} = 0 & 0 < z < 18 \\ u|_{z=0} = w|_{z=0} = 5.88 & 0.3 < z < 0.7 \\ p|_{z=18} = 0 \\ u = v = w = 0 & \text{other BCs} \end{cases} \quad (26)$$

In Eq. (26), density $\rho = 1.1614\,\text{kg/m}^3$, kinematic viscosity $\nu = 1.5895\,\text{m}^2/\text{s}$, the units of velocity ($u$, $v$, $w$), pressure ($p$) and coordinate ($z$, $r$) are m/s, Pa, mm, respectively.

In this numerical test, a mesh resolution of 128×1536 is specified for both input and output. The input is the initial field which is filtered from upstream to downstream

$$I(n+1) = I(n) \otimes [1/3 \quad 1/3 \quad 1/3] \quad n = 0,1,2,\cdots N_x(\text{or } N_x) \quad (27)$$

where *n* is the distance from the upstream boundary to an internal row or column.

The receptive field sizes of the six Encoder-Decoder pairs are 129, 65, 33, 17, 9, and 5, respectively, which are obtained by dilating a 3×3 convolution kernel, and the relevant parameters are shown in Table 4. The padded virtual nodes are obtained by fitting second-order Taylor polynomials for implementing the 8-order central difference. The loss function contains the PDE loss, the boundary condition loss, and the data loss. The known data here is given by two measured velocity components *u* and *v*. The data that needs to be inferred are the unknown vertical velocity component *w* and the pressure *p*. In this way, we make up for the inability of the PIV (Particle Image Velocimetry) experiment to measure the vertical velocity component and pressure. The loss weights are set to be 16384 for mass and momentum PDE losses in the *z*, *r*, *θ* directions, and to be 1 for the boundary condition and data losses. During the solving process of MRF-PINN, the Adam optimiser is used to optimise 30000 epochs with a learning rate of 1e-4. The L-BFGS optimiser is employed to fine-tune MRF-PINN for 1000 epochs.

Table 4

Details of the MRF-PINN dilated convolution for solving Navier-Stokes PDEs

| Enc-Dec | Enc-Dec-1 | Enc-Dec-2 | Enc-Dec-3 | Enc-Dec-4 | Enc-Dec-5 | Enc-Dec-6 |
| --- | --- | --- | --- | --- | --- | --- |
| $k$ | 2 | 4 | 8 | 16 | 32 | 64 |
| Dilation rate | 64 | 32 | 16 | 8 | 4 | 2 |
| Padding | 64 | 32 | 16 | 8 | 4 | 2 |
| Receptive field | 129 | 65 | 33 | 17 | 9 | 5 |
| Receptive field/min(h,w) | 129/128≈1.0 | 65/128≈0.51 | 33/128≈0.26 | 17/128≈0.13 | 9/128≈0.09 | 5/128≈0.04 |

For FVM, the mesh resolution is 150×1800. A second-order discretization is used for pressure and the momentum equation is discretised by a second-order QUICK scheme.

Figs. 9-10 demonstrate the validity of MRF-PINN which can capture the solutions by FVM well. The solving errors $\varepsilon$ of the predictions by MRF-PINN for $u$, $v$, $w$ and $p$ are 0.0609, 0.0249, 0.0376, and 0.0522, respectively.

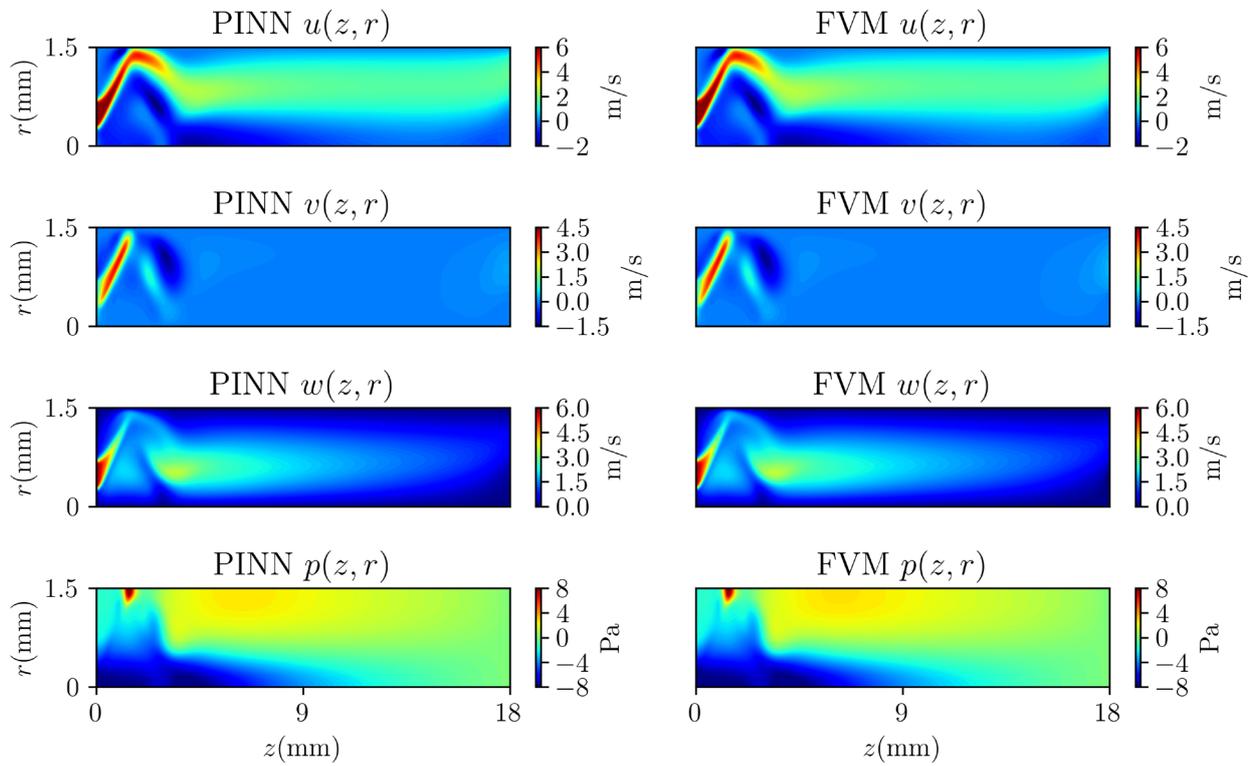

Fig. 9. Solutions of the Navier-Stokes PDEs by MRF-PINN and FVM

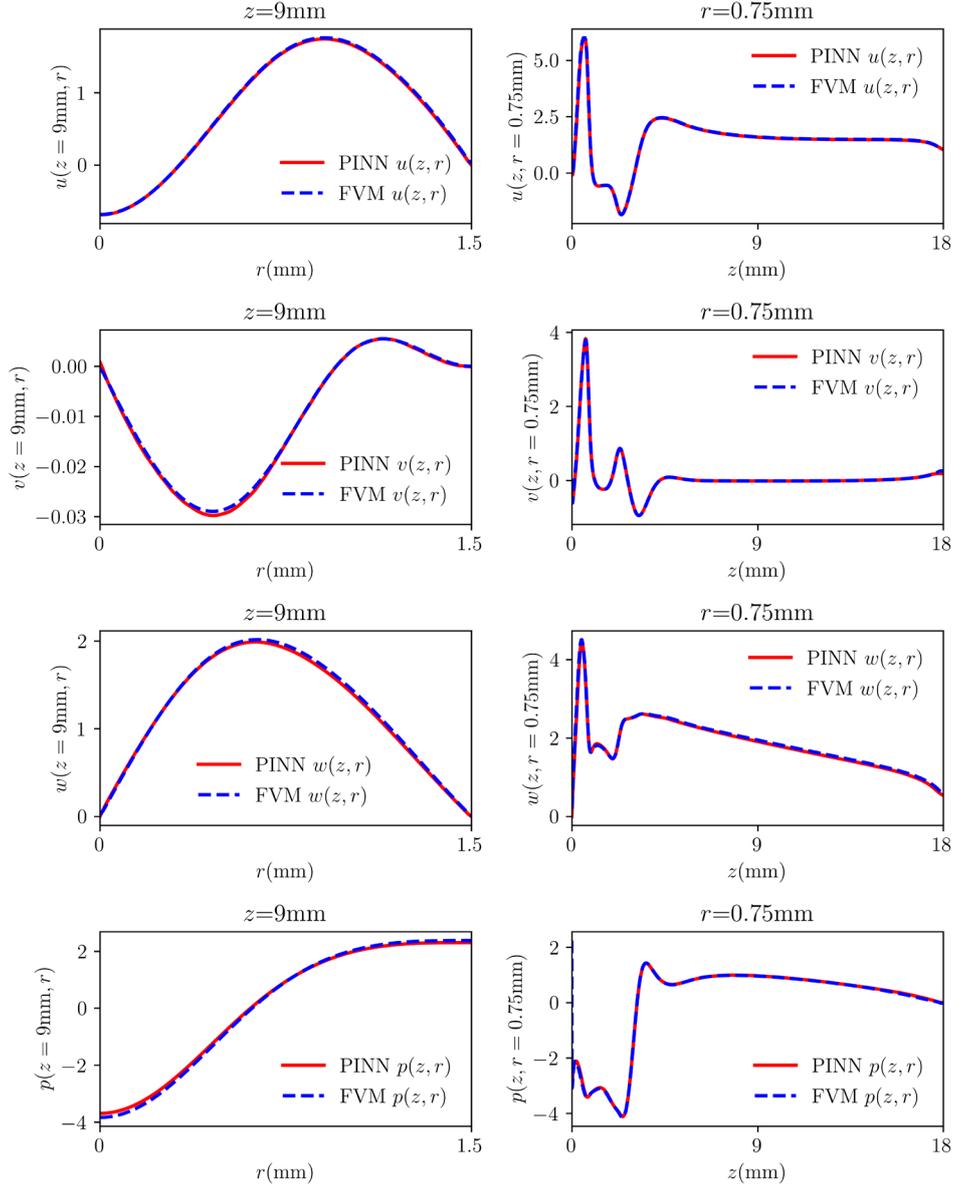

Fig. 10. Solutions of the Navier-Stokes PDEs by MRF-PINN and FVM

### 3.3 Mesh resolution

With traditional numerical calculation methods such as FEM and FVM, the mesh resolution will directly affect the solving error and cost. In the validation of MRF-PINN, it is found that MRF-PINN can also effectively reduce the solving error under high mesh resolution. However, the FLOPs and the mesh number increase proportionally.

As shown in Tables 5-7 for solving linear PDEs, when the mesh resolution is increased from 32×64 to 128×256 (the mesh number is increased by 16 times), the

FLOPs also increase by 16 times, and the solving errors are reduced by 79%, 12%, and 66%, respectively(Except for parabolic PDE at mesh resolution 128×256).

As shown in Table 8 for solving the nonlinear Navier-Stokes PDEs, increasing the mesh resolution from 32 × 64 to 128 × 256, the fitting errors of $u$ and $v$ are reduced by 55% and 74%, respectively, and the inference errors of $w$ and $p$ are reduced by 59% and 49%. This proves that high-resolution mesh can suppress the solving error in MRF-PINN, but the FLOPs increase simultaneously.

Table 5

Elliptic PDE: the influence of mesh resolution on the solving error and cost

| Mesh resolution | 32×64 | 64×128 | 128×256 |
|---|---|---|---|
| Solving error | 0.0127 | 0.00621 | **0.00263** |
| FLOPs | **770M** | 3078M | 12312M |

Table 6

Parabolic PDE: the influence of mesh resolution on the solving error and cost

| Mesh resolution | 32×64 | 64×128 | 128×256 |
|---|---|---|---|
| Solving error | 0.00736 | **0.00645** | 0.0113 |
| FLOPs | **770M** | 3078M | 12312M |

Table 7

Hyperbolic PDE: the influence of mesh resolution on the solving error and cost

| Mesh resolution | 32×64 | 64×128 | 128×256 |
|---|---|---|---|
| Solving error | 0.0551 | 0.0326 | **0.0187** |
| FLOPs | **770M** | 3078M | 12312M |

Table 8

Navier-Stokes PDEs: the influence of mesh resolution on the solving error and cost

| Mesh resolution | 32×384 | 64×768 | 128×1536 |
|---|---|---|---|
| Solving error of $u$ | 0.135 | 0.0861 | **0.0609** |
| Solving error of $v$ | 0.0947 | 0.0627 | **0.0249** |
| Solving error of w | 0.0927 | 0.0572 | **0.0376** |
| Solving error of p | 0.103 | 0.0726 | **0.0522** |
| FLOPs | **5127M** | 20509M | 82035M |

## 4. Algorithm advantages

Based on the above MRF-PINN, FEM, and FVM algorithms and settings, the advantages of MRF-PINN, the dimensional balance method and high-order finite difference will be discussed in this section.

**4.1 MRF-PINN**

This paper proposes the resolution-adapted MRF-PINN to avoid the time-consuming hyperparameter tuning. A correlation coefficient $R$ is defined to support quantifying the contribution of the Encoder-Decoder pairs with different receptive fields to the final MRF-PINN result.

$$R_i = \frac{|u_i \cdot u|}{\|u_i\|_2 \|u\|_2} \qquad (28)$$

As $R$ varies from 0 to 1, the similarity between the Encoder-Decoder output and the final MRF-PINN result is increased.

The effect of the channel number on the $R$ value is tested for each PDE and the results are shown in Figs. 11-14, and $u_1$, …, $u_6$ are the outputs of Encoder-Decoder-1, …, Encoder-Decoder-6, respectively. As shown in Tables 3-4, the size of the receptive field decreases from Encoder-Decoder-1 to Encoder-Decoder-6. From an overall perspective, there is no strong regularity among the $R$ value, the channel number, and the receptive field. This indicates that PINN does not tend to converge to a predictable receptive field, however, it chooses the most reasonable receptive field composition. Therefore, a fixed receptive field cannot ensure the generality of the PINN. To avoid the costly manual tunning, the multi-receptive field approach is proposed and its advantages are corroborated by the following tests.

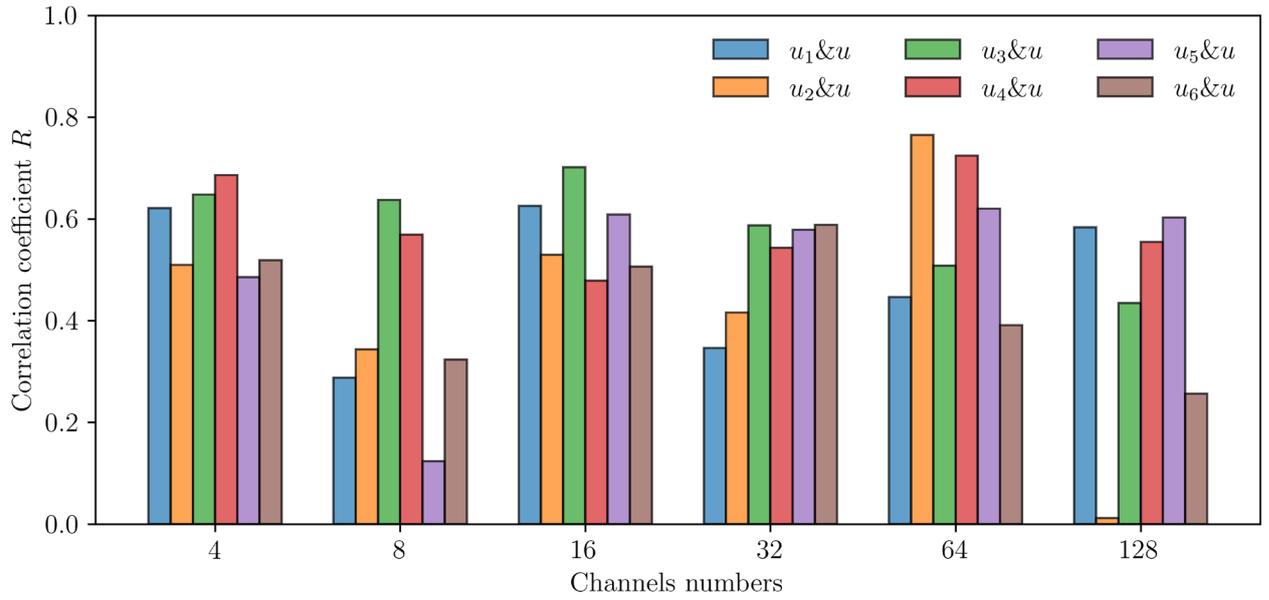

Fig. 11. Correlation coefficients($R$) for elliptic equations.

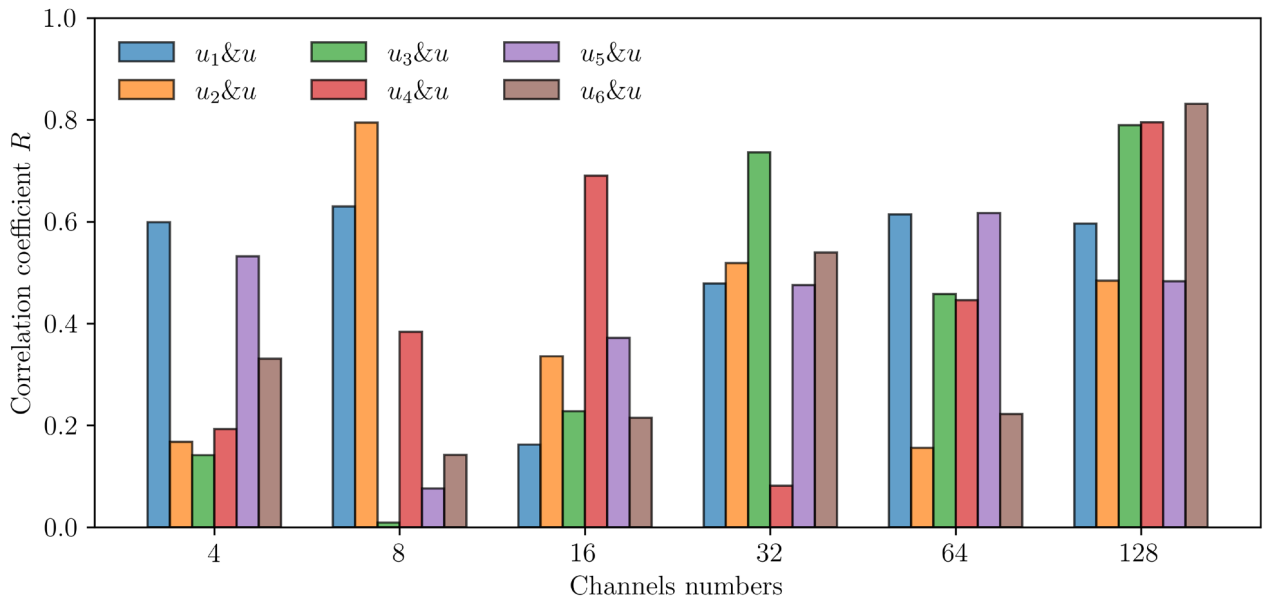

Fig. 12. Correlation coefficients($R$) for parabolic equations.

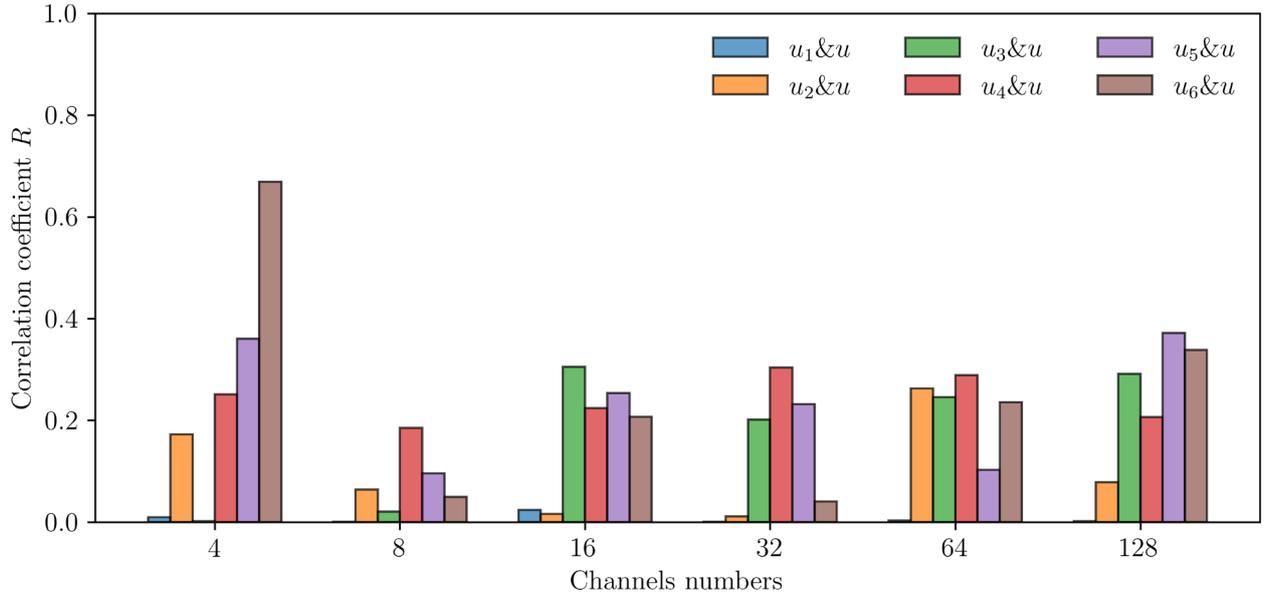

Fig. 13. Correlation coefficients($R$) for hyperbolic equations.

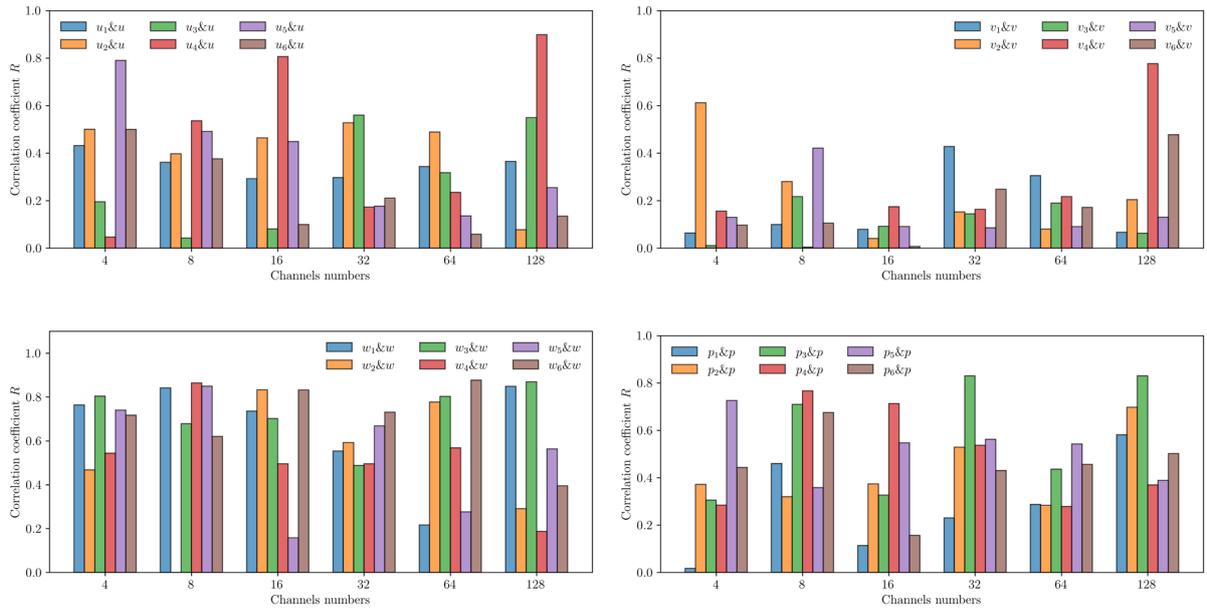

Fig. 14. Correlation coefficient($R$) for the Navier-Stokes PDEs.

We test PINN with five fixed receptive fields (3, 5, 7, 9, and 11) where their solving errors are compared with that of the MRF-PINN. The mesh resolutions are 128×256 for solving linear PDEs and 128×1536 for Navier-Stokes PDEs. As shown in Tables 9-12, the optimal fixed receptive fields for the ellipse, parabola, hyperbolic, and Navier-Stokes PDEs are 3, 7, 11, and 11, respectively. Overall, the solving error of MRF-PINN is comparable to the optimal fixed receptive field and the error of MRF-PINN is 3.5% larger, 3.4% smaller, and 20% larger for the linear PDEs. Compared with the optimal fixed receptive fields solving the nonlinear Navier-Stokes PDEs, the fitting errors for $u$ and $v$ of MRF-PINN are 0.16% larger and 9.1% smaller, the inference errors for $w$ and $p$ of MRF-PINN are 1.5% and 1.1% smaller, respectively. These comparisons can prove that the resolution-adapted MRF-PINN can reduce the tuning cost and suppress the solving error well.

Table 9

Errors of solving elliptic PDE using fixed receptive field PINN and MRF-PINN

| Receptive field | 3 | 5 | 7 | 9 | 11 | MRF-PINN |
|---|---|---|---|---|---|---|
| Solving error | **0.00254** | 0.00528 | 0.00290 | 0.00290 | 0.00295 | 0.00263(Top2) |

Table 10

Errors of solving parabolic PDE using fixed receptive field PINN and MRF-PINN

| receptive field | 3 | 5 | 7 | 9 | 11 | MRF-PINN |
|---|---|---|---|---|---|---|
| Solving error | 0.0318 | 0.0191 | 0.0117 | 0.0198 | 0.0192 | **0.0113** |

Table 11

Errors of solving hyperbolic PDE using fixed receptive field PINN and MRF-PINN

| Receptive field | 3 | 5 | 7 | 9 | 11 | MRF-PINN |
|---|---|---|---|---|---|---|
| Solving error | 0.0162 | 0.0184 | 0.0257 | 0.0248 | **0.0155** | 0.0187 |

Table 12

Errors of solving Navier-Stokes PDEs using fixed receptive field PINN and MRF-PINN

| Receptive field | 3 | 5 | 7 | 9 | 11 | MRF-PINN |
|---|---|---|---|---|---|---|
| Solving error of $u$ | 0.0613 | **0.0608** | 0.0612 | **0.0608** | 0.0611 | 0.0609 |
| Solving error of $v$ | 0.0337 | 0.0451 | 0.0475 | 0.0456 | 0.0274 | **0.0249** |
| Solving error of $w$ | 0.0404 | 0.0587 | 0.0467 | 0.0463 | 0.0382 | **0.0376** |
| Solving error of $p$ | 0.0532 | 0.0615 | 0.0539 | 0.0563 | 0.0528 | **0.0522** |

## 4.2 Dimensional balance method

As mentioned in Section 2.3, the dimensional balance method is used to estimate the loss weights when solving the Navier-Stokes PDEs. Its advantages to accelerate the optimisation and convergence are demonstrated here.

The dimensional balance method only performs back-propagation once in each epoch, while the dynamic weight method requires $l+1$ back-propagations ($l$ is the number of loss functions). As shown in Fig. 15, compared with the dynamic weight method, the dimensional balance method leads to the reduction of the computational time by 95% and 72% for Adam and L-BFGS, respectively.

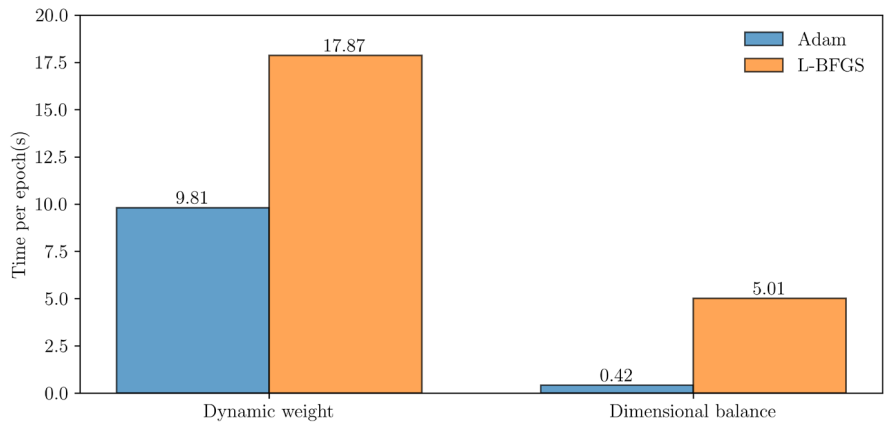

Fig. 15. Solving time of the dynamic weight method and dimensional balance method in one epoch

Figs. 16-17 show that, compared with the dynamic weight method, the dimensional balance method leads to a more stable solving process as expected until the convergence is reached.

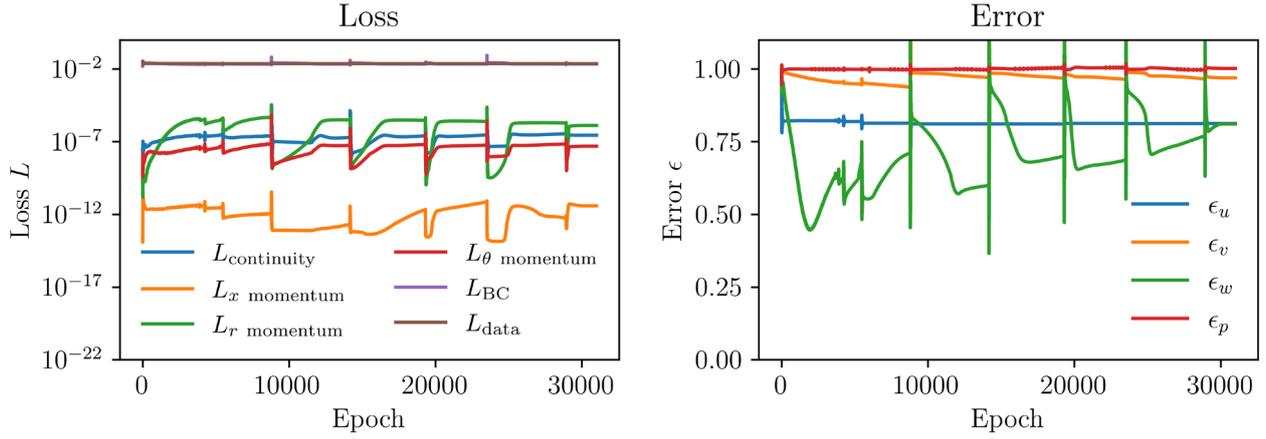

Fig. 16. Solving loss and error when using the dynamic weight method

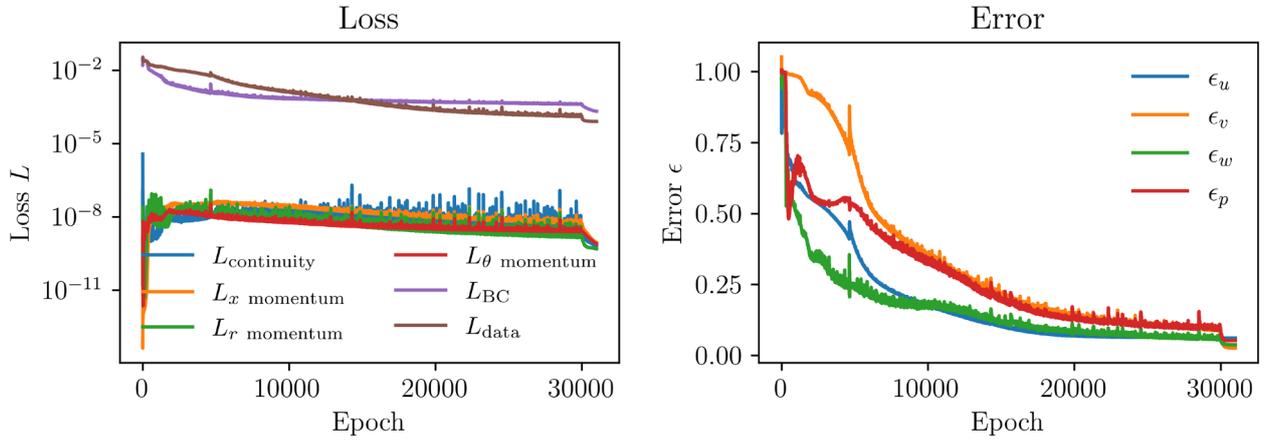

Fig. 17. Solving loss and error when using the dimensional balance method

### 4.3 Finite difference accuracy

#### 4.3.1 Elliptic PDE

As shown in Table 13, a high-order finite difference can lead to the reduction of the MRF-PINN solving error for the elliptic PDE overall. The exception occurs for the channel number of 128 where the solving error of the 6-order difference is the lowest but is very close to the accuracy of the 8-order difference. In addition, the solving error does not show a significant dependence on the channel number.

Table 13

Elliptic PDE: the influence of the channel number and the finite difference accuracy on the solving error and cost

| Channel number | 2-order difference | 4-order difference | 6-order difference | 8-order difference | FLOPs | Parameters |
|---|---|---|---|---|---|---|
| 4 | 0.0129 | 0.0127 | 0.0127 | **0.0125** | 2.43M | 4.95K |
| 32 | 0.0129 | **0.0127** | **0.0127** | **0.0127** | 58.4M | 281K |
| 128 | 0.0129 | 0.0128 | **0.0126** | 0.0127 | 769M | 4442K |

**4.3.2 Parabolic PDE**

As shown in Table 14, for the parabolic PDE the solving errors of the MRF-PINN are at a low level which is smaller than 0.01 independent of the channel number and the finite difference. Overall, high-order difference has limited improvement. For 32 and 128 channels, the 8-order difference acquires the lowest error.

Table 14

Parabolic PDE: the influence of the channel number and the finite difference accuracy on the solving error and cost

| Channel number | 2-order difference | 4-order difference | 6-order difference | 8-order difference | FLOPs | Parameters |
|---|---|---|---|---|---|---|
| 4 | 0.00758 | **0.00748** | 0.00783 | 0.00823 | 2.43M | 4.95K |
| 32 | 0.00769 | 0.00709 | 0.00734 | **0.00707** | 58.4M | 281K |
| 128 | 0.00781 | 0.00758 | 0.00787 | **0.00736** | 769M | 4442K |

**4.3.3 Hyperbolic PDE**

As shown in Table 15, the higher-order finite difference significantly suppresses the solving error for solving the hyperbolic equation. Compared with the 2-order finite difference, the solving errors of the 8-order difference for the three channel numbers decrease by 15%, 4.3%, and 4.5%, respectively. More channels tend to reduce the solving error.

Table 15

Hyperbolic PDE: the influence of the channel number and the finite difference accuracy on the solving error and cost

| Channel number | 2-order difference | 4-order difference | 6-order difference | 8-order difference | FLOPs | Parameters |
|---|---|---|---|---|---|---|
| 4 | 0.0654 | **0.0554** | 0.0609 | **0.0554** | 2.43M | 4.95K |
| 32 | 0.0576 | 0.0555 | 0.0554 | **0.0551** | 58.4M | 281K |
| 128 | 0.0577 | 0.0554 | 0.0554 | **0.0551** | 769M | 4442K |

**4.3.4 Navier-Stokes PDEs**

As mentioned in Section 3.2, the fitting and the inference errors are discussed for the MRF-PINN model for solving Navier-Stokes PDEs. As shown in Tables 16-17, the fitting error of the velocity component $u$ is hardly affected by the finite difference accuracy and the channel number. The fitting error of the velocity component $v$ tends to decrease for more channels and higher order finite difference. Since the velocity components $u$ and $v$ are directly supervised by the known data, they are mainly affected by the constraint of data loss.

As shown in Tables 18-19, the solving error is generally reduced as the finite difference accuracy is improved. The low-order finite difference can actually make the inference error of $w$ lower when the channel number of the MRF-PINN is low. Moreover, as the channel number increases, the higher-order finite difference remakes the inference error of $w$ higher than that of the lower-order finite difference. The inference error of $p$ is similar to the conclusion of $w$. More channels and the high-order finite difference can reduce the inference error effectively. Compared with 4 channels and 2-order finite difference, the inference errors of $w$ and $p$ with 128 channels and 8-order difference are reduced by 69% and 77%, respectively. At the same time, if we adopt the high channel number and high-order finite difference, the fitting error of the velocity components $u$ and $v$ does not have much influence.

Table 16
Fitting error of $u$

| Channel number | 2-order difference | 4-order difference | 6-order difference | 8-order difference | FLOPs | Parameters |
|---|---|---|---|---|---|---|
| 4 | 0.133 | **0.132** | 0.133 | 0.133 | 30.9M | 6.28K |
| 32 | **0.133** | 0.134 | 0.134 | 0.134 | 478M | 291K |
| 128 | **0.134** | **0.134** | **0.134** | 0.135 | 5127M | 4484K |

Table 17

Fitting error of $v$

| Channel number | 2-order difference | 4-order difference | 6-order difference | 8-order difference | FLOPs | Parameters |
|---|---|---|---|---|---|---|
| 4 | **0.102** | 0.133 | 0.147 | 0.136 | 30.9M | 6.28K |
| 32 | 0.105 | 0.0975 | 0.0963 | **0.094** | 478M | 291K |
| 128 | 0.105 | 0.0965 | 0.0955 | **0.0947** | 5127M | 4484K |

Table 18

Inference error of $w$

| Channel number | 2-order difference | 4-order difference | 6-order difference | 8-order difference | FLOPs | Parameters |
|---|---|---|---|---|---|---|
| 4 | **0.295** | 0.361 | 0.313 | 0.375 | 30.9M | 6.28K |
| 32 | 0.0970 | **0.0921** | 0.0934 | 0.0948 | 478M | 291K |
| 128 | 0.0976 | 0.0938 | **0.0920** | 0.0927 | 5127M | 4484K |

Table 19

Inference error of $p$

| Channel number | 2-order difference | 4-order difference | 6-order difference | 8-order difference | FLOPs | Parameters |
|---|---|---|---|---|---|---|
| 4 | 0.413 | 0.537 | **0.366** | 0.462 | 30.9M | 6.28K |
| 32 | 0.137 | 0.115 | **0.114** | 0.121 | 478M | 291K |
| 128 | 0.141 | 0.107 | 0.105 | **0.103** | 5127M | 4484K |

## 5. Summary

This paper proposes a multiple receptive field convolutional PINN(MRF-PINN) model adaptive to different equation types and mesh resolution. The MRF-PINN can solve multiple types of partial differential equations (PDEs) with low solving error and no hyperparameter tuning. In order to improve the calculation accuracy of PDE loss, we adopt the Taylor polynomial to pad the boundary of the PINN result so that it can perform high-order finite difference. In the case that the Navier-Stokes equations have many loss functions and significant difference in loss magnitudes, we first use the dimensional balance method to estimate the loss weights.

Compared with the traditional fixed receptive field PINN, the MRF-PINN has the advantages of low manual tuning cost and low solving error. When solving Navier-Stokes equations, our proposed dimensional balance method has a shorter solving time and better convergence than the dynamic weight method. Based on MRF-PINN, we systematically study the effects of finite difference accuracy, model complexity (channel number), and mesh resolution on PINN results. Generally, similar to the traditional numerical solving method (such as FEM and FVM), the high-order finite difference is beneficial to obtaining low-error PDE solutions. Besides, the high-order finite difference does not bring about numerical instability in our tests. Higher model complexity and mesh resolution can reduce the error of the PDE solution but will simultaneously increase the FLOPs cost.

This work implies that the MRF-PINN model can adapt to different PDE types, which is expected to be a general scheme for convolutional PINN. Researchers can also use the dimensional balance method to estimate the loss weights to solve the Navier-Stokes equations. Finally, systematic numerical experiments on finite difference, model complexity, and mesh resolution fill the current weak research on convolutional PINN. Researchers are expected to achieve low-error solving results and good numerical stability at the high-order finite difference and high mesh resolution.

## CRediT authorship contribution statement

**Shihong Zhang:** Conceptualization, Methodology, Software, Validation, Writing - Original Draft, Writing - Review & Editing. **Bosen Wang:** Supervision, Writing - Review & Editing. **Chi Zhang:** Supervision.

**Declaration of competing interest**

The authors declare that they have no known competing financial interests or personal relationships that could have appeared to influence the work reported in this paper.

**Acknowledgements**

Thanks to the National Natural Science Foundation of China (52106128) and the National Science and Technology Major Project (J2019-III-0014-0057) for their financial support.